\documentclass[nohyperref]{article}

\usepackage{microtype}
\usepackage{graphicx}
\usepackage{booktabs} %

\usepackage[backref=page]{hyperref}

\usepackage[accepted]{icml2022}

\usepackage{amsmath}
\usepackage{amssymb}
\usepackage{mathtools}
\usepackage{amsthm}

\theoremstyle{plain}
\newtheorem{thm}{Theorem}[section]

\newtheorem{lem*}[thm]{Lemma}
\newtheorem{cor}[thm]{Corollary}
\theoremstyle{definition}
\newtheorem{dfn}[thm]{Definition}

\theoremstyle{remark}

\usepackage{color}
\usepackage{bm}
\usepackage{subcaption}
\usepackage{wrapfig}
\usepackage{mathabx} %
\newcommand{\BB}{\mathbb{B}}

\newcommand{\RR}{\mathbb{R}}
\newcommand{\Sph}{\mathbb{S}}
\newcommand{\HH}{\mathbb{H}}
\newcommand{\NN}{\mathbb{N}}

\newcommand{\PP}{\mathbb{P}}

\newcommand{\CC}{\mathbb{C}}

\newcommand{\dd}{\mathrm{d}}
\newcommand{\eps}{\varepsilon}

\newcommand{\Sch}{\mathcal{S}}

\newcommand{\xx}{{\bm{x}}}
\newcommand{\yy}{{\bm{y}}}
\newcommand{\xxi}{{\bm{\xi}}}
\newcommand{\oo}{{\bm{o}}}

\newcommand{\llambda}{{\bm{\lambda}}}
\renewcommand{\aa}{{\bm{a}}}

\newcommand{\cc}{{\bm{c}}}
\newcommand{\uu}{{\bm{u}}}
\newcommand{\vv}{{\bm{v}}}

\renewcommand{\ss}{{\bm{s}}}

\newcommand{\rrho}{{\bm{\varrho}}}

\newcommand{\iprod}[1]{\langle#1\rangle}

\newcommand{\wdot}{\,\cdot\,}

\newcommand{\lieG}{\mathfrak{g}}
\newcommand{\lieK}{\mathfrak{k}}
\newcommand{\lieA}{\mathfrak{a}}
\newcommand{\lieN}{\mathfrak{n}}
\newcommand{\bdX}{\partial X}

\newcommand{\eprod}[1]{(#1)}
\newcommand{\bdD}{\partial D}

\newcommand{\bdHH}{\partial\HH}

\newcommand{\Hp}{\mathbb{B}}
\newcommand{\bdHp}{\partial\Hp}
\newcommand{\spd}{\PP}
\newcommand{\bdspd}{\partial\spd}

\newcommand{\cdom}{Z}

\DeclareMathOperator{\sign}{sign}

\DeclareMathOperator{\tr}{tr}

\DeclareMathOperator{\vol}{vol}
\DeclareMathOperator{\lip}{Lip}
\DeclareMathOperator{\ind}{\mathbf{1}}

\newcommand{\iiprod}[1]{(\!(#1)\!)}

\newcommand{\reffig}[1]{Figure~\ref{fig:#1}}
\newcommand{\reftab}[1]{Table~\ref{tab:#1}}
\newcommand{\refapp}[1]{Appendix~\ref{sec:#1}}
\newcommand{\refsec}[1]{\S~\ref{sec:#1}}

\newcommand{\refthm}[1]{Theorem~\ref{thm:#1}}

\allowdisplaybreaks

\icmltitlerunning{Fully-Connected Network on Noncompact Symmetric Space}

\begin{document}

\twocolumn[
\icmltitle{Fully-Connected Network on Noncompact Symmetric Space \\ and Ridgelet Transform based on Helgason-Fourier Analysis}

\icmlsetsymbol{equal}{*}

\begin{icmlauthorlist}
\icmlauthor{Sho Sonoda}{riken}
\icmlauthor{Isao Ishikawa}{ehime,riken}
\icmlauthor{Masahiro Ikeda}{riken}
\end{icmlauthorlist}

\icmlaffiliation{riken}{RIKEN Center for Advanced Intelligence Project (AIP), Tokyo, Japan}
\icmlaffiliation{ehime}{Ehime University, Ehime, Japan}

\icmlcorrespondingauthor{Sho Sonoda}{sho.sonoda@riken.jp}

\icmlkeywords{neural network, ridgelet transform, noncompact symmetric space, Helgason-Fourier analysis}

\vskip 0.3in
]

\printAffiliationsAndNotice{}  %

\begin{abstract}
Neural network on Riemannian symmetric space such as hyperbolic space and the manifold of symmetric positive definite (SPD) matrices is an emerging subject of research in geometric deep learning. Based on the well-established framework of the Helgason-Fourier transform on the noncompact symmetric space, we present a fully-connected network and its associated ridgelet transform on the noncompact symmetric space, covering the hyperbolic neural network (HNN) and the SPDNet as special cases. The ridgelet transform is an analysis operator of a depth-2 continuous network spanned by neurons, namely, it maps an arbitrary given function to the weights of a network. Thanks to the coordinate-free reformulation, the role of nonlinear activation functions is revealed to be a wavelet function. Moreover, the reconstruction formula is applied to present a constructive proof of the universality of finite networks on symmetric spaces.
\end{abstract}

\section{Introduction}

Geometric deep learning is an emerging research direction that aims to devise neural networks on non-Euclidean spaces \citep{Bronstein2021}. %
In this study, we focus on devising a fully-connected layer on a noncompact symmetric space $X=G/K$ \citep{Helgason.GGA,Helgason.GASS}. 
In general, it is more challenging to devise a fully-connected layer on a manifold 
than to devise a convolution layer because neither the scalar product, bias translation, nor pointwise activation can be trivially defined. A noncompact symmetric space is a Riemannian manifold $X$ with nonpositive curvature, as well as a homogeneous space $G/K$ of Lie groups $G$ and $K$. It covers several important spaces in the recent literature of representation learning, such as the hyperbolic space and the manifold of symmetric positive definite (SPD) matrices, or the SPD manifold. On those spaces, several neural networks have been developed such as \emph{hyperbolic neural networks (HNNs)} %
and \emph{SPDNets}. %

\paragraph{Neural Network on Hyperbolic Space.}
The hyperbolic space is a symmetric space with a constant negative curvature.
Following the success of Poincar\'{e} embedding \citep{Krioukov2010,Nickel2017,Nickel2018,Sala2018},
the hyperbolic space has been recognized as an effective space for embedding tree-structured data;
and hyperbolic neural networks (HNNs) \citep{Ganea2018hnn,Gulcehre2019,Shimizu2021} have been developed to promote effective use of hyperbolic geometry for saving parameters against Euclidean counterparts. 
The previous studies such as HNN \citep{Ganea2018hnn} and HNN++ \citep{Shimizu2021} have replaced each operation with gyrovector calculus, but there are still rooms for arguments such as on the expressive power of the proposed network and on the role of nonlinear activation functions. 

\paragraph{Neural Network on SPD Manifold.}
The SPD manifold equipped with the standard Riemannian metric
has nonconstant nor nonpositive curvature. 
The metric is isomorphic to the Fisher information metric for multivariate centered normal distributions. 
Since covariance matrices are positive definite, the SPD manifold has been investigated and applied in a longer and wider literature than the hyperbolic space. Besides, the SPD manifold has also attracted attention as a space for graph embedding \citep{Lopez2021,Cruceru2021}. 
To reduce the computational cost without harming the Riemannian geometry, several distances have been proposed such as the affine-invariant Riemannian metric (AIRM) \citep{Pennec2006}, the Stein metric \citep{Sra2012}, the Bures–Wasserstein metric \citep{Bhatia2019}, the Log-Euclidean metric \citep{Arsigny2006,Arsigny2007}, and the vector-valued distance \citep{Lopez2021}. Furthermore, neural networks on SPD manifolds have been developed, such as SPDNet \citep{Huang2017,Dong2017,Gao2019,Brooks2019,Brooks2019a}, deep manifold-to-manifold transforming network (DMT-Net) \citep{Zhang2018b}, and ManifoldNet \citep{Chakraborty2018,Chakraborty2022}.

Although those networks are aware of underlying geometry, except for a universality result on \emph{horospherical} HNNs by \citet{Wang2021}, previous studies lack theoretical investigations, such as on the expressive power and on the effect of nonlinear activation functions. The purpose of this study is to define a fully-connected layer on a noncompact symmetric space in a unified manner from the perspective of harmonic analysis on symmetric space, %
and derive an associated \emph{ridgelet transform}---an analysis operator that maps a function $f$ on $X$ to the weight parameters, written $\gamma$, of a network. 
In the end, the ridgelet transform is given as a \emph{closed-form expression}, the reconstruction formula further elicits a constructive proof of the \emph{universality} of finite models, and the role/effect of an activation function will be understood as a wavelet function.

\paragraph{Harmonic Analysis on Symmetric Space.}
The Helgason-Fourier transform has been introduced in \citep{Helgason1965} as a Fourier transform on the noncompact symmetric space $X$. This is an integral transform of functions $f$ on $X$ with respect to the eigenfunctions of the Laplace-Beltrami operator $\Delta_X$ on $X$. We refer to \citet[Introduction]{Helgason.GGA} and \citet[Ch.III]{Helgason.GASS} for more details.

\paragraph{The Integral Representation $S[\gamma](\xx)$ on Euclidean Space} is an infinite-dimensional linear representation of a depth-2 fully-connected neural network, given by the following integral operator: For every $\xx \in \RR^m$,
\begin{align}
S[\gamma](\xx) = \int_{\RR^m\times\RR} \gamma(\aa,b)\sigma(\aa\cdot\xx-b)\dd\aa\dd b. %
\label{eq:conti.enn}
\end{align}
Here, each function $\xx\mapsto\sigma(\aa\cdot\xx-b)$ represents a single neuron, or a feature map of input $\xx$ parametrized by $(\aa,b)$.
The integration over $(\aa,b)$ implies that all the possible neurons are assigned in advance, and thus $S[\gamma]$ can be understood as a \emph{continuous neural network}. 
We note, however, that if we take $\gamma$ to be a finite sum of Dirac's measures such as $\gamma_p := \sum_{i=1}^p c_i \delta_{(\aa_i,b_i)}$, then the integral representation can also exactly reproduce a finite model: For every $\xx \in \RR^m$,
\begin{align*}
S[\gamma_p](\xx) = \sum_{i=1}^p c_i \sigma(\aa_i\cdot\xx-b_i).
\end{align*}
In summary, $S[\gamma]$ is a mathematical model of shallow neural networks with \emph{any} width ranging from finite to infinite.

\paragraph{The Ridgelet Transform $R[f;\rho](\aa,b)$} is a right inverse (or pseudo-inverse) operator of the integral representation operator $S$. For the Euclidean neural network given in \eqref{eq:conti.enn}, the ridgelet transform is given as a \emph{closed-form expression}: For every $(\aa,b) \in \RR^m\times\RR$,
\begin{align}
    R[f;\rho](\aa,b) := \int_{\RR^m} f(\xx) \overline{\rho(\aa\cdot\xx-b)}\dd\xx. %
    \label{eq:ridgelet.enn}
\end{align}
Here, $f:\RR^m\to\CC$ is a target function to be approximated, and $\rho:\RR\to\CC$ is an auxiliary function, called the \emph{ridgelet function}. 
Under mild conditions, the reconstruction formula
\begin{align*}
    S[R[f;\rho]] = \iiprod{\sigma,\rho} f,
\end{align*}
holds, where $\iiprod{\cdot,\cdot}$ denote a scalar product of $\sigma$ and $\rho$ given by a weighted inner-product in the Fourier domain as
\begin{align*}
    \iiprod{\sigma,\rho} := (2\pi)^{m-1}\int_\RR \sigma^\sharp(\omega)\overline{\rho^\sharp(\omega)}|\omega|^{-m}\dd\omega,
\end{align*}
where $\cdot^\sharp$ denotes the Fourier transform in $b \in \RR$. Therefore, as long as the product $\iiprod{\sigma,\rho}$ is neither $0$ nor $\infty$, we can normalize $\rho$ to satisfy $\iiprod{\sigma,\rho}=1$ so that $S[R[f;\rho]]=f$. 

In other words, $R$ and $S$ are analysis and synthesis operators, and thus play the same roles as the Fourier ($F$) and inverse Fourier ($F^{-1}$) transforms respectively, in the sense that the reconstruction formula $S[R[f;\rho]]=\iiprod{\sigma,\rho}f$ corresponds to the Fourier inversion formula $F^{-1}[F[f]]=f$. 
In the meanwhile, different from the case of the Fourier transform, there are infinitely many different $\rho$'s satisfying $\iiprod{\sigma,\rho}=1$. This means that $R$ is not strictly an inverse operator to $S$, which is unique if it exists, but a right inverse operator, indicating that $S$ has a nontrivial null space. \citet{Sonoda2021ghost} have revealed that the null space is spanned by the ridgelet transforms $R[\wdot;\rho_0]$ with degenerated ridgelet functions satisfying $\iiprod{\sigma,\rho_0}=0$. This means that any parameter distribution $\gamma$ satisfying $S[\gamma]=f$ can always be represented as (not always single but) a linear combination of ridgelet transforms.

Despite the common belief that neural network parameters are a blackbox, the closed-form expression of ridgelet transform \eqref{eq:ridgelet.enn} clearly describes how the network parameters are organized, which is a clear advantage of the integral representation theory. Furthermore,
the integral representation theory can deal with a wide range of activation functions without any modification, not only ReLU but all the tempered distribution $\Sch'(\RR)$ \citep[see, e.g.,][]{Sonoda2015acha}.

\paragraph{Relations between Continuous and Finite Models.}
(1) The relation between a general continuous model $\int \gamma(v)\sigma(v,x)\dd v$ with a general feature map $x \mapsto \sigma(v,x)$ parametrized by $v$, and the finite model $\sum_{i=1}^p c_i \sigma(v_i,x)$ is well investigated in the  \emph{Maurey-Jones-Barron (MJB) theory}, claiming the $L^p$-density of finite models in the space of continuous models \citep[see, e.g.,][]{kainen.survey}. The density is fundamental to show that a certain property of finite models is preserved when the model is extended to a continuous model, so that we can concentrate on investigating the continuous model instead of the finite model. 
(2) In addition, \citet{Sonoda2020aistats} have shown that the parameter distribution of a finite model trained by regularized empirical risk minimization (RERM) converges to a certain unique ridgelet spectrum $R[f;\sigma_*]$ with special $\sigma_*$ in an over-parametrized regime. This means that we can understand the parameters at local minima to be a finite approximation of the ridgelet transform, and thus we can investigate the ridgelet transform to study the minimizer of the learning problem.

\paragraph{Historical Overview.}
The idea of the integral representation first emerged in the 1990s to investigate the expressive power of infinitely-wide shallow neural networks \citep{Irie1988,Funahashi1989,Carroll.Dickinson,Ito.Radon,Barron1993}, and the original ridgelet transform is discovered  independently by \citet{Murata1996}, \citet{Candes.PhD} and \citet{Rubin.calderon}. In the context of sparse signal processing, ridgelet analysis has been developed as a multidimensional counterpart of wavelet analysis \citep{Donoho2002,Starck2010,Kutyniok2012,Kostadinova2014}. In the context of deep learning theory, continuous models have been employed in the so-called \emph{mean-field theory} to show the global convergence of the SGD training of shallow ReLU networks \citep{Nitanda2017,Mei2018,Rotskoff2018,Chizat2018,Sirignano2020,Suzuki2020}, and new \emph{ridgelet transforms for ReLU networks} have been developed to investigate the expressive power of ReLU networks \citep{Sonoda2015acha},
and to establish the \emph{representer theorem} for ReLU networks \citep{Savarese2019,Ongie2020,Parhi2020,Unser2019}. %

\subsection*{Contributions of This Study}
One of the major shortcomings of conventional ridgelet analysis has been that the closed-form expression like \eqref{eq:ridgelet.enn} is known only for the case of Euclidean network: $\sigma(\aa\cdot\xx-b)$. 
In this study, we explain a natural way to find the ridgelet transform via the Fourier expression, then obtain a series of new ridgelet transforms for noncompact symmetric space $X=G/K$ in a unified manner by replacing the Euclidean-Fourier transform with the Helgason-Fourier transform on noncompact symmetric space.
The reconstruction formula $S[R[f]]=f$ can provide a constructive proof of the universal approximation property of finite neural networks on an arbitrary noncompact symmetric space. As far as we have noticed, \citet{Wang2021} is the only author who shows the universality of HNNs. Following the classical arguments by \citet{Cybenko1989}, her proof is based on the Hahn-Banach theorem. As a result, it is simple but non-constructive.
On the other hand, our results (1) are more informative because of the constructive nature, 
(2) cover a wider range of spaces, i.e., any noncompact symmetric space $X=G/K$, and (3) cover a wider range of nonlinear activation functions, i.e., any tempered distribution $\sigma \in \Sch'(\RR)$, without any modification. %

\paragraph{Remarks for Avoiding Potential Confusions.} As clarified in the discussion, the HNNs devised in this study is the same as the one investigated by \citet{Wang2021}, but different from the ones proposed by \citet{Ganea2018hnn} and \citet{Shimizu2021}. Both HNNs can be regarded as extensions of the Euclidean NN (ENN), but as a hyperbolic counterpart of the Euclidean hyperplane, \citet{Wang2021} and we employed the \emph{horosphere}, while \citet{Ganea2018hnn} and \citet{Shimizu2021} employed the \emph{set of geodesics}, called the \emph{Poincar\'e hyperplane}. As a consequence, our main results \emph{do} cover our \emph{horospherical HNN}, but do not cover their \emph{geodesic HNNs}. Yet, we conjecture that the proof technique can be applied for those geodesic HNNs. On the other hand, the SPDNet devised in this study is essentially the same as the ones proposed in previous studies \citep{Huang2017,Dong2017,Gao2019,Brooks2019,Brooks2019a}.

We further remark that the so-called ``equivalence between convolutional networks and fully-connected networks'' \citep[e.g.][]{Petersen2020} can hold \emph{only when} networks are carefully designed (e.g., when $X$ is a finite set). 
While a convolution on $X$ is a binary operation $f * g$ of functions $f,g:X\to\RR$, a scalar-product on $X$ is a binary operation $\iprod{x,y}$ of points $x,y \in X$, and there are \emph{no} canonical rules to identify functions and points in general. 
Moreover, there are \emph{no} canonical scalar-products for points in general manifolds. 
Therefore, even if we are given a convolutional network on $X$, we cannot directly translate it as a fully-connected network.

\subsection*{Notations}
$|\cdot|_E$ (or simply $|\cdot|$) denotes the Euclidean norm of $\RR^m$. %

$GL(m,\RR)$ denotes the set of $m \times m$ real regular matrices, or the general linear group. $O(m)$ denotes the set of $m \times m$ orthogonal matrices, or the orthogonal group. $D(m), D_+(m)$ and $D_{\pm 1}(m)$ denote the sets of $m \times m$ diagonal matrices with real entries, positive entries, and $\pm 1$, respectively. $T_+(m), T_1(m)$ and $T_0(m)$ denote the sets of $m \times m$ upper triangular matrices with positive entries, ones, and zeros on the diagonal, respectively. $\spd_m$ denotes the set of $m \times m$ symmetric positive definite matrices.

On a (possibly noncompact) manifold $X$, $C_c(X)$ denotes the compactly supported continuous functions, $C_c^\infty(X)$ denotes the compactly supported infinitely differentiable functions, and $L^2(X)$ denotes the square-integrable functions. When $X$ is a symmetric space, $\dd x$ is supposed to be the left-invariant measure.

For any integer $d>0$, $\Sch(\RR^d)$ denotes the Schwartz test functions (or rapidly decreasing functions) on $\RR^d$, and $\Sch'(\RR^d)$ the tempered distributions on $\RR^d$ (i.e., the topological dual of $\Sch(\RR^d)$). We eventually set the class of activation functions to be tempered distributions $\Sch'(\RR)$, which covers truncated power functions $\sigma(b)=b_+^k = \max\{b,0\}^k$ covering the step function for $k=0$ and the rectified linear unit (ReLU) for $k=1$. We refer to \citet{Grafakos.classic,GelfandShilov,Sonoda2015acha} for more details on Schwartz distributions and Fourier analysis on them.

To avoid potential confusion, we use two symbols $\widehat{\cdot}$ and $\cdot^\sharp$ for the Fourier transforms in the input variable $x \in X=G/K$ (or $\xx \in \RR^m$) and the bias variable $b \in \RR$, respectively. For example,
    $\widehat{f}(\xxi) := \int_{\RR^m}f(\xx)e^{-i\xx\cdot\xxi}\dd\xx$ for $\xxi \in \RR^m$, 
    $\rho^\sharp(\omega) := \int_\RR \rho(b)e^{-ib\omega}\dd b$ for $\omega \in \RR$, and 
    $\gamma^\sharp(\aa,\omega) = \int_\RR \gamma(\aa,b)e^{-ib\omega}\dd b$ for $(\aa,\omega) \in \RR^m\times\RR$.

\section{Fully-Connected Layer on Euclidean Space}
We briefly review the Euclidean fully-connected layer $\xx\mapsto\sigma(\aa\cdot\xx-b)$ on $\RR^m$, where $\xx \in \RR^m$ is an input vector, $(\aa,b) \in \RR^m\times\RR$ is hidden parameters, $\aa\cdot\xx$ is the Euclidean scalar product, and $\sigma:\RR\to\RR$ is an arbitrary given nonlinear function. In particular, expressions \eqref{eq:coordinate.free} and \eqref{eq:fourier.enn} are keys to devise fully connected layer on a symmetric space with a variety of activation function $\sigma$ and to derive the associated ridgelet transform.

\subsection{Coordinate-Free Reformulation of Euclidean Fully-Connected Layer} \label{sec:geometric.fc.e}

\begin{figure}
    \centering
    \includegraphics[width=.8\columnwidth]{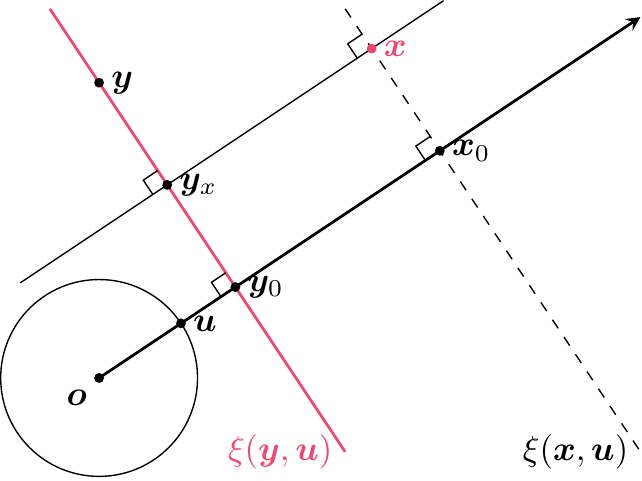} %
    \caption{The Euclidean fully-connected layer $\sigma(\aa\cdot\xx-b)$ is recast as the signed distance $d(\xx,\xi)$ from a point $\xx$ to a hyperplane $\xi(\yy,\uu)$ followed by a wavelet function $\sigma(r\cdot)$, where $\yy$ satisfies $r\yy\cdot\uu=b$ and $\xi(\yy,\uu)$ passes through the point $\yy$ with a normal $\uu$.}
    \label{fig:euclid}
\end{figure}

For any $(\xx,\uu) \in \RR^m\times\Sph^{m-1}$, put
\begin{align*}
    \quad \xi(\xx,\uu) := \{ \yy \in \RR^m \mid \uu\cdot(\xx-\yy)=0 \},
\end{align*}
the hyperplane passing through a point $\xx \in \RR^m$ and orthogonal to a unit vector $\uu \in \Sph^{m-1}$.

First, we change the parameters in polar coordinates as
\begin{align*}
    (\aa,b) = (r \uu, r\uu \cdot \yy), \quad (r,\uu,\yy) \in \RR_{\ge 0} \times\Sph^{m-1}\times\RR^m.
\end{align*}
We note that the mapping from $\yy$ to $b$ is not injective, but it is rather understood as the mapping from (any representative point of) hyperplane $\xi((b/r)\uu,\uu) = \{ \yy  \mid  r\uu\cdot\yy=b \}$ to $b$.

Then, the fully-connected layer $\sigma(\aa\cdot\xx-b)$ is rewritten as
\begin{align}
    \sigma(\aa\cdot\xx-b)
    &= \sigma(r\uu\cdot(\xx-\yy)) %
    = \sigma(r d(\xx,\yy_x)) \notag \\
    &= \sigma(r d(\xx,\xi(\yy,\uu))), \label{eq:coordinate.free}
\end{align}
where $d(\xx,\yy) := \sign(\xx-\yy)|\xx-\yy|_E$ denotes the signed Euclidean distance, %
and $\yy_x$ denotes the closest point to $\xx$ in the hyperplane $\{\yy\mid r\uu\cdot\yy=b\}$. \reffig{euclid} illustrates the relations of symbols.

The last two expressions are coordinate-free, but the final expression \eqref{eq:coordinate.free} is much appreciated because $\yy$ can be an arbitrary representative point of hyperplane $\{r\uu\cdot\yy=b\}$. Meanwhile, the scaled nonlinear function $\sigma(r \cdot)$ is understood as a wavelet function, which plays a role of multiscale analysis \citep{Mallat2009book} such as singularity detection with scale $r$ running from $0$ to $\infty$.

In summary, a fully-connected layer $\sigma(\aa\cdot\xx-b)$ is recast as $\sigma(r d(\xx,\xi))$ ``wavelet analysis with respect to a wavelet function $\sigma$ on the signed distance $d(\xx,\xi)$ between point $\xx$ and hyperplane $\xi$.'' Since the wavelet transform can detect a point singularity, wavelet analysis on the distance between a point and a hyperplane can detect a singularity in the normal direction to the hyperplane.

\subsection{How to Solve $S[\gamma]=f$ and Find $\gamma=R[f]$} \label{sec:solve}

We explain the basic steps to find the parameter distribution $\gamma$ satisfying $S[\gamma]=f$. The basic steps is three-fold: (Step~1) Turn the network into the \emph{Fourier expression}, (Step~2) \emph{change variables} to split the feature map into useful and junk factors, and (Step~3) put the unknown $\gamma$ the \emph{separation-of-variables form} to find a particular solution.

The following procedure is valid, for example, when $\sigma \in \Sch'(\RR), \rho \in \Sch(\RR), f \in L^2(\RR^m)$ and $\gamma \in L^2(\RR^m\times\RR)$. See \citet{Kostadinova2014} and \citet{Sonoda2015acha} for more details on the valid combinations of function classes.
\paragraph{Step~1.}
To begin with, we turn the network into the \emph{Fourier expression} as below.
\begin{align*}
    S[\gamma](\xx)
    &:= \int_{\RR^m\times\RR} \gamma(\aa,b) \sigma( \aa\cdot\xx-b )\dd\aa\dd b \notag \\
    &= \int_{\RR^m} [\gamma(\aa,\cdot) *_b \sigma](\aa\cdot\xx)\dd\aa \notag \\
    &= \frac{1}{2\pi} \int_{\RR^m\times\RR} \gamma^\sharp(\aa,\omega) \sigma^\sharp( \omega ) e^{i\omega \aa\cdot \xx}\dd\aa\dd \omega.
    \end{align*}
    Here, 
    $*_b$ denotes the convolution in $b$; and the last equation follows from the identity (i.e., the Fourier inversion formula) $\phi(b) = \frac{1}{2\pi}\int_\RR \phi^\sharp(\omega)e^{i\omega b}\dd\omega$ with $\phi(b) = [\gamma(\aa,\cdot)*_b\sigma](b)$ and $b = \aa\cdot\xx$.
\paragraph{Step~2.} Next, we change variables $(\aa,\omega) = (\xxi/\omega,\omega)$ with $\dd \aa \dd \omega = |\omega|^{-m} \dd \xxi\dd\omega$ so that the modified feature map $\sigma^\sharp(\omega)e^{i\omega\aa\cdot\xx}$ split into useful and junk factors as
    \begin{align}
    &= \frac{1}{2\pi} \int_\RR \left[ \int_{\RR^m} \gamma^\sharp(\xxi/\omega,\omega) e^{i\xxi\cdot \xx}\dd\xxi \right] 
    \sigma^\sharp( \omega ) |\omega|^{-m} \dd \omega. \label{eq:fourier.enn}
\end{align}
\paragraph{Step~3.} Finally, since inside the bracket $[\cdots]$ is the Fourier inversion with respect to $\xxi$, it is natural to put $\gamma$ to be a \emph{separation-of-variables} expression
\begin{align}
    \gamma_{f,\rho}^\sharp(\xxi/\omega, \omega) := \widehat{f}(\xxi) \overline{\rho^\sharp(\omega)}, \label{eq:sep.var}
\end{align}
with the given function $f \in L^2(\RR^m)$ and an arbitrary function $\rho \in \Sch(\RR)$.
Then, we have
\begin{align*}
    S[\gamma_{f,\rho}](\xx)
    &= \iiprod{\sigma,\rho} \frac{1}{(2\pi)^m}\int_{\RR^m} \widehat{f}(\xxi) e^{i\xxi\cdot \xx}\dd\xxi\dd \omega \notag \\
    &= \iiprod{\sigma,\rho} f(\xx),
\end{align*}
where we put
\begin{align*}
    \iiprod{\sigma,\rho} := (2\pi)^{m-1}\int_\RR \sigma^\sharp(\omega)\overline{\rho^\sharp(\omega)}|\omega|^{-m}\dd\omega.
\end{align*}
In other words, the separation-of-variables expression $\gamma_{f,\rho}$ is a particular solution to the integral equation $S[\gamma]=cf$ with factor $c = \iiprod{\sigma,\rho} \in \CC$.

In the end, $\gamma_{f,\rho}$ turns out to be the ridgelet transform: The Fourier inversion of 
    $\gamma_{f,\rho}^\sharp(\aa,\omega) = \widehat{f}(\omega\aa) \overline{\rho^\sharp(\omega)}$
is calculated as
\begin{align*}
    \gamma_{f,\rho}(\aa,b)
    &= \frac{1}{2\pi}\int_{\RR} \widehat{f}(\omega\aa) \overline{\rho^\sharp(\omega) e^{-i\omega b}}\dd\omega \notag \\
    &= \frac{1}{2\pi}\int_{\RR^m \times \RR} f(\xx) \overline{\rho^\sharp(\omega) e^{i \omega (\aa\cdot\xx - b)}}\dd\omega \notag \\
    &= \int_{\RR^m \times \RR} f(\xx) \overline{\rho(\aa\cdot\xx-b)} \dd\xx,
\end{align*}
which is exactly the definition of the ridgelet transform $R[f;\rho]$.

In conclusion, the separation-of-variables expression \eqref{eq:sep.var} is the way to naturally find the ridgelet transform.
We note that Steps~1 and 2 to obtain \eqref{eq:fourier.enn} can be understood as the \emph{change-of-frame} from the neurons $\sigma(\aa\cdot\xx-b)\dd\aa\dd b$, which we are less familiar with, to the tensor product of a plane wave and a junk: $e^{i\xxi\cdot\xx}\dd\xxi \otimes \sigma^\sharp(\omega)|\omega|^{-m}\dd\omega$, which we are much familiar with. Hence, the map $\gamma(\aa,b) \mapsto \gamma^\sharp(\xxi/\omega,\omega)$ is understood to be the associated transformation of the coefficient $\gamma$.

\section{Harmonic Analysis on Noncompact Symmetric Space}

Readers may skip the first subsection, \refsec{x}, by understanding general notations in symmetric space $X$ as specific ones in the hyperbolic space or the SPD manifold as listed in \reftab{corresp}.
\begin{table*}[t]
    \centering
    \begin{tabular}{lll}
    \toprule
        in symmetric space & in hyperbolic space & in SPD manifold\\
    \midrule
        $X=G/K$ & hyperbolic space $\HH^m$ & SPD manifold $\spd_m$ \\
        $\bdX := K/M$ & boundary (or ideal sphere) $\bdHH^m$ & boundary $\bdspd_m$ \\
        $\lieA^*$ & frequency domain $\RR^1$ & frequency domain $\RR^m$ \\
        $\xi(x,u)$ & horosphere $\xi(x,u)$ & horosphere $\xi(x,u)$ \\ %
        $\iprod{x,u} := -H(g^{-1}k)$ & signed distance $\iprod{x,u}$ & vector-valued distance $\iprod{x,u}$ \\
    \bottomrule
    \end{tabular}
    \caption{Correspondence of notations in $X=G/K, \HH^m$ and $\spd_m$}
    \label{tab:corresp}
\end{table*}

\subsection{Noncompact Riemannian Symmetric Space} \label{sec:x}

We follow the notation by \citet[Ch.~II]{Helgason.GASS} except for the conflict cases. For example, we assign ``$u \in \partial X$'' instead of ``$b \in B$'' for the boundary of $X$, since $b$ is assigned for the bias in a fully-connected layer in this study.

Let $G$ be a connected semisimple Lie group with finite center, and let $G=KAN$ be its Iwasawa decomposition. Namely, it is a unique diffeomorphic decomposition of $G$ into subgroups $K,A,$ and $N$, where $K$ is maximal compact, $A$ is maximal abelian, and $N$ is maximal nilpotent. 
For example, when $G=GL(m,\RR)$ (general linear group), then $K=O(m)$ (orthogonal group), $A = D_+(m)$ (all positive diagonal matrices), and $N = T_1(m)$ (all upper triangular matrices with ones on the diagonal).

Let $\dd g, \dd k, \dd a,$ and $\dd n$ be left $G$-invariant measures on $G,K,A,$ and $N$ respectively. Following \citet[Proposition~5.1, Ch.~I]{Helgason.GGA}, we normalize the measures so that $\int_K \dd k = 1$, and 
\begin{align*}
    \int_G f(g)\dd g
    &= \int_{KAN} f(kan) e^{2 \varrho \log a} \dd k \dd a \dd n \notag \\
    &= \int_{NAK} f(nak) e^{-2 \varrho \log a} \dd n \dd a \dd k \notag \\
    &= \int_{ANK} f(ank) \dd a \dd n \dd k,
\end{align*}
for any $f \in C_c(G)$, with a constant $\varrho \in \lieA^*$ defined below. 

Let $\lieG, \lieK,\lieA,$ and $\lieN$ be the Lie algebras of $G,K,A,$ and $N$ respectively.
By a fundamental property of abelian Lie algebra, both $\lieA$ and its dual $\lieA^*$ are the same dimensional vector spaces, and thus they can be identified with $\RR^r$ for some $r$, namely $\lieA = \lieA^* = \RR^r$. We call $r := \dim \lieA$ the rank of $X$. For example, when $G=GL(m,\RR)$, then $\lieG=\mathfrak{gl}_m = \RR^{m\times m}$ (all $m \times m$ real matrices), $\lieK = \mathfrak{o}_m$ (all skew-symmetric matrices), $\lieA=D(m)$ (all diagonal matrices), and $\lieN=T_0(m)$ (all strictly upper triangular matrices).

Let $X := G/K$ be a noncompact symmetric space, namely, a Riemannian manifold composed of all the left cosets
\begin{align*}
X := G/K := \{ x=gK \mid g \in G \}.
\end{align*}
Using the identity element $e$ of $G$, let $o=eK$ be the origin of $X$. By the construction of $X$, group $G$ acts transitively on $X$, and let $g[x] := ghK$ (for $x=hK$) denote the $G$-action of $g \in G$ on $X$. Specifically, any point $x\in X$ can always be written as $x=g[o]$ for some $g \in G$. Let $\dd x$ denote the left $G$-invariant measure on $X$. Following the normalization above, we normalize $\dd x$ so that $\int_X f(x) \dd x := \int_G f(g[o]) \dd g = \int_{AN} f(an[o])\dd a \dd n$ for any $f \in C_c(X)$.

Let $M := C_K(A) := \{ k \in K \mid ka = ak \mbox{ for all } a \in A \}$ be the centralizer of $A$ in $K$, and let 
\begin{align*}
\bdX := K/M := \{ u=kM \mid k \in K \}
\end{align*}
be the boundary (or ideal sphere) of $X$, which is known to be a compact manifold. Let $\dd u$ denote the uniform probability measure on $\bdX$.
For example, when $K=O(m)$ and $A=D_+(m)$, then $M = D_{\pm 1}$ (the subgroup of $K$ consisting of diagonal matrices with entries $\pm 1$).

Let
\begin{align*}
\Xi := G/MN := \{ \xi = gMN \mid g \in G\} %
\end{align*}
be the space of horospheres. 
Here, basic horospheres are: An $N$-orbit $\xi_o := N[o] = \{ n[o] \mid n \in N \}$, which is a horosphere passing through the origin $x=o$ with normal $u=eM$;
and $ka[\xi_o] = kaN[o]$, which is a horosphere through point $x=ka[o]$ with normal $u=kM$. In fact, any horosphere can be represented as $\xi(kan[o], kM)$ since $kaN = kanN$ for any $n \in N$.
We refer to \citet[Ch.I, \S~1]{Helgason.GASS} and \citet[\S~3.5]{Bartolucci2021} for more details on the horospheres and boudaries.

As a consequence of the Iwasawa decomposition, for any $g \in G$ there uniquely exists an $r$-dimensional vector $H(g) \in \lieA$ satisfying $g \in K e^{H(g)}N$. For any $(x,u) = (g[o], kM) \in X \times \bdX$, put 
\begin{align*}
\iprod{x,u} := -H(g^{-1}k) \in \lieA \cong \RR^r,
\end{align*}
which is understood as the $r$-dimensional vector-valued distance, called the \emph{composite distance}, from the origin $o\in X$ to the horosphere $\xi(x,u)$ through point $x$ with normal $u$. Here, the vector-valued distance means that the $\ell^2$-norm coincides with the Riemannian length, that is, $|\iprod{x,u}| = |d( o, \xi(x,u) )|$.
We refer to \citet[Ch.II, \S~1, 4]{Helgason.GASS} and \citet[\S~2]{Kapovich2017} for more details on the vector-valued composite distance.

Let $\Sigma \subset \lieA^*$ be the set of (restricted) roots of $\lieG$ with respect to $\lieA$. For $\alpha \in \Sigma$, let $\lieG_\alpha$ denote the corresponding root space, and call $m_\alpha := \dim (\lieG_\alpha)$ the multiplicity of $\alpha$. Let $\lieA^+$ be the Weyl chamber corresponding to $\lieN$, i.e. $\lieN = \sum_{\alpha\in\Sigma^+}\lieG_\alpha$, where $\Sigma^+$ is the set of $\alpha\in\Sigma$ that are positive on $\lieA^+$. Put 
$\varrho := \sum_{\alpha\in\Sigma^+}\frac{m_\alpha}{2} \alpha  \in \lieA^*.$
Let $W$ be the Weyl group of $G/K$, and let $|W|$ denote its order. Let $\cc(\lambda)$ 
be the Harish-Chandra $\cc$-function for $G$. We refer to \citet[Theorem~6.14, Ch.~IV]{Helgason.GGA} for the closed-form expression of the $\cc$-function.

\subsection{Helgason-Fourier Transform on Symmetric Space}
For any function $f$ on $X$, the Helgason-Fourier transform is defined as
\begin{align*}
    \widehat{f}(\lambda,u) &:= \int_{X} f(x) e^{(-i\lambda+\varrho)\iprod{x,u}}\dd x, \ (\lambda,u) \in \lieA^* \times \bdX
\end{align*}
where the exponent $(-i\lambda+\varrho)\iprod{x,u}$ is understood as the action of functional $-i\lambda+\varrho \in \lieA^*$ on a vector $\iprod{x,u} \in \lieA$. The inversion formula \citep[Theorems~1.3 and 1.5, Ch.~III]{Helgason.GASS} is given by
\begin{align*}
f(x) &= \int_{\lieA^*\times\bdX} \widehat{f}(\lambda,u) e^{(i\lambda+\varrho)\iprod{x,u}}\frac{\dd\lambda\dd u}{|W||\cc(\lambda)|^{2}}. %
\end{align*}
Here, the equality holds at every point $x \in X$ when $f \in C_c^\infty(X)$, and in $L^2$ when $f \in L^2(X)$. 
In particular, the following Plancherel theorem holds: For any $f_1,f_2 \in L^2(X)$,
    $\int_X f_1(x) \overline{f_2(x)} \dd x = \int_{\lieA^* \times \bdX} \widehat{f_1}(\lambda,u) \overline{\widehat{f_2}(\lambda,u)}\frac{\dd\lambda\dd u}{|W||\cc(\lambda)|^2}.$

The integral kernel $e^{(-i\lambda+\varrho)\iprod{x,u}}$ is an $X$-counterpart of a plane wave $e^{-i\lambda\uu\cdot\xx}$ in the Euclidean-Fourier transform $\int_{\RR^m} f(\xx) e^{-i\lambda\uu\cdot\xx}\dd\xx$ (expressed in polar coordinate).
While the plane wave $e^{-i\lambda\uu\cdot\xx}$ is a joint eigenfunction of all the invariant differential operators (that is, all the polynomials of the Laplacian $\varDelta$) on $\RR^m$, the $X$-plane wave $e^{(-i\lambda+\varrho)\iprod{x,u}}$ is a joint eigenfunction of all the invariant differential operators (e.g., polynomials of the Laplace-Beltrami operator $\varDelta_X$) on $X$. 
In particular, the Plancherel measure $|\cc(\lambda)|^{-2}\dd\lambda\dd u$ plays a parallel role to $\lambda^{-m}\dd \lambda\dd\uu$ in polar coordinates.

\subsection{Poincar\'{e} Ball Model of Hyperbolic Space}
\begin{figure}
    \centering
    \includegraphics[width=.8\columnwidth]{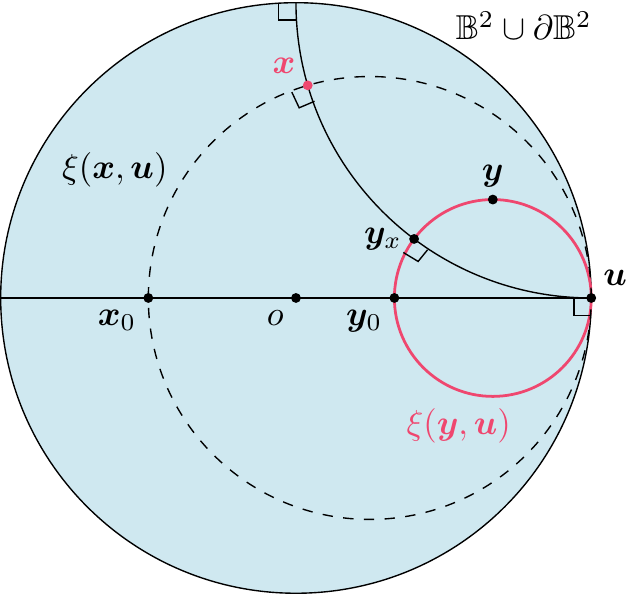} %
    \caption{Poincar\'{e} disk $\BB^2$, boundary $\bdHp^2$, point $\xx$ (magenta), horocycle $\xi(\yy,\uu)$ (magenta) through point $\yy$ tangent to the boundary at $\uu$, and two geodesics (solid black)  orthogonal to the boundary at $\uu$ through $\oo$ and $\xx$ respectively. The signed composite distance $\iprod{\yy,\uu}$ from the origin $\oo$ to the horocycle $\xi(\yy,\uu)$ can be visualized as the Riemannian distance from $\oo$ to point $\yy_0$. Similarly, the distance between point $\xx$ and horocycle $\xi(\yy,\uu)$ is understood as the Riemannian distance between $\xx$ and $\yy_x$ along the geodesic, or equivalently, $\xx_0$ and $\yy_0$.}
    \label{fig:poincare}
\end{figure}
Here, we briefly introduce the Poincar\'{e} ball $\Hp^m$ as a Riemannian manifold. In \refapp{poincare.ss}, we further explain the homogeneous space aspect of the Poincar\'{e} disk $\Hp^2$. In the following, the boldface such as $\xx$ and $\uu$ emphasizes that the symbols should be understood as the Cartesian coordinates, rather than a point itself on a manifold.

Let $\Hp^m := \{\xx \in \RR^m \mid |\xx|_E<1\}$ be a Riemannian manifold equipped with metric
\begin{align*}
    \mathfrak{g}_{\xx} := \left(\frac{2}{1-|\xx|_E^2}\right)^2\sum_{i=1}^m \dd x_i \wedge \dd x_i, \quad \xx \in \Hp^m.
\end{align*} This is the Poincar\'e ball model of $m$-dimensional hyperbolic space $\HH^m$. The Riemannian distance between $\xx,\yy \in \Hp^m$ is given by
\begin{align*}
d_P(\xx,\yy)=\cosh^{-1}\left(1+\frac{2|\xx-\yy|_E^2}{(1-|\xx|_E^2)(1-|\yy|_E^2)} \right),
\end{align*}
and the Riemannian volume measure at $\xx\in\Hp^m$ is given by 
\begin{align*}
\dd\vol_{\mathfrak{g}}(\xx) =
\left(\frac{2}{1-|\xx|_E^2}\right)^m\dd\xx, %
\end{align*}
with respect to the Lebesgue measure $\dd\xx$.
Let $\bdHp^m := \{ \uu \in \RR^m \mid |\uu|_E=1 \} = \Sph^{m-1}$ be the boundary (or ideal sphere) equipped with the uniform spherical measure $\dd\uu$. Since $\HH^m$ is rank-one, we identify $\lieA^* \cong \RR^1$ equipped with the Lebesgue measure.

In the Poincar\'e ball model $\Hp^m$, any boundary point $\uu$ on the boundary $\bdHp^m$ is infinitely far from any inner point $\xx$ in $\Hp^m$;
any geodesic is a Euclidean arc that is orthogonal to the boundary $\bdHp^m$;
any hyperbolic ball/sphere is a Euclidean ball/sphere in $\Hp^m$;
and any horosphere is a Euclidean ball that is tangent to the boundary $\bdHp^m$.
Hence a horosphere is understood as a ``hyperbolic sphere of infinite radius'', and it is identified by two parameters $(\xx,\uu) \in \Hp^m \times\bdHp^m$ as ``a horosphere $\xi(\xx,\uu)$ passing through $\xx$ tangent to the boundary at $\uu$.''
We note that since a hyperplane in the Euclidean space can also be understood as a ``Euclidean sphere of infinite radius'', we can understand horospheres as a hyperbolic counterpart of hyperplanes in the Euclidean space.

The signed composite distance $\iprod{\xx,\uu}$ from the origin $\bm{o}$ to the horosphere $\xi(\xx,\uu)$ is calculated as
\begin{align*}
    \iprod{\xx,\uu} := d_P(\bm{o},\xi(\xx,\uu)) = d_P(\bm{o},\xx_0) = \log \left( \frac{1-|\xx|_E^2}{|\xx-\uu|_E^2} \right).
\end{align*}
Here, we put $\xx_0 := t \uu$ for some $|t| < 1$ so that $(\xx_0 - \xx,\uu-\xx)_E=0$, i.e., Thales' theorem.

The Helgason-Fourier transform and the inversion formula are instantiated as
\begin{align*}
    &\widehat{f}(\lambda,\uu) = \int_{\Hp^m} f(\xx) e^{(-i\lambda+\varrho)\iprod{\xx,\uu}} \left(\frac{2}{1-|\xx|_E^2}\right)^m \dd \xx,\\
    &f(\xx) = \frac{c_m^2}{2}\int_{\RR\times\Sph^{m-1}}\widehat{f}(\lambda,\uu)e^{(i\lambda+\varrho)\iprod{\xx,\uu}}\frac{\dd\lambda\dd\uu}{|\cc(\lambda)|^{2}},
\end{align*}
for any $(\lambda,\uu) \in \RR\times\Sph^{m-1}$ and $\xx \in \Hp^m$ respectively,
where $c_m^2 = 2^{2\varrho} / (2\pi \vol(\Sph^{m-1})), \varrho = (m-1)/2,$ 
and the Plancherel measure $|\cc(\lambda)|^{-2}$ is given by
\begin{align*}
    ( 2^{k-1} (2k-1)!! )^{-2} \prod_{j=0}^{k-1}(\lambda^2+j^2),
\end{align*}
when $m=2k+1$, and 
\begin{align*}
    ( 2^{k-1} (2k-2)!! )^{-2} \frac{\pi\lambda \tanh(\pi\lambda)}{\lambda^2 + (1/2)^2} %
    \prod_{j=0}^{k-1}\left({\textstyle \lambda^2+\left(\frac{2j-1}{2}\right)^2}\right),
\end{align*}
when $m=2k$.

\section{Fully-Connected Layer on Symmetric Space}

We define the fully-connected layer on the noncompact symmetric space, present the associated ridgelet transform and reconstruction formula, and finally state the $cc$-universality of finite networks.

\subsection{Network Definition}

In accordance with the geometric perspective, it is natural to define the network as below.
\begin{dfn} %
Let $\sigma:\RR\to\CC$ be a measurable function. For any function $\gamma:\lieA^*\times\bdX\times\RR\to\CC$, the continuous neural network on the symmetric space $X$ is given by
\begin{align*}
    &S[\gamma](x)
    \\
    &:= \int_{\lieA^*\times\bdX\times\RR} \gamma(a,u,b)\sigma( a \iprod{x,u} - b ) e^{\varrho \iprod{x,u}} \dd a \dd u \dd b.
\end{align*}
Here, we call $x \in X$ the input, $a \in \lieA^*$ the scale, $u \in \bdX$ the normal (of horosphere), and $b \in \RR$ the bias. $\varrho \in \lieA^*$ is a constant vector depending on $G/K$. 
\end{dfn}
If we take $y \in X$ satisfying $a\iprod{y,u}=b$, then we can rewrite $a\iprod{x,u}-b$ as $a d(x,\xi(y,u))$, which can be understood as an $X$-counterpart of the coordinate-free expression \eqref{eq:coordinate.free}.
For technical reasons (i.e., for connecting the Helgason-Fourier transform), we impose an auxiliary weight $e^{\varrho\iprod{x,u}}$.

\subsection{Ridgelet Transform}
\begin{dfn} %
Let $\rho:\RR\to\CC$ and $f:X\to\CC$ be measurable functions. Put
\begin{align*}
    &R[f;\rho](a,u,b) := \int_X \cc[f](x)\overline{\rho(a\iprod{x,u}-b)}e^{\varrho\iprod{x,u}}\dd x,\\
    &\cc[f](x) := \int_{\lieA^*\times\bdX}\widehat{f}(\lambda,u)e^{(i\lambda+\varrho)\iprod{x,u}}\frac{\dd\lambda\dd u}{|W||\cc(\lambda)|^{4}},\\
    &\iiprod{\sigma,\rho} := \frac{|W|}{2\pi}\int_\RR \sigma^\sharp(\omega)\overline{\rho^\sharp(\omega)}|\omega|^{-r}\dd\omega.
\end{align*}
Here $\cc[f]$ is defined as a multiplier satisfying $\widehat{\cc[f]}(\lambda,u)=\widehat{f}(\lambda,u)|\cc(\lambda)|^{-2}$.
\end{dfn}

\subsection{Reconstruction Formula}
\begin{thm}[Reconstruction Formula on Symmetric Space] \label{thm:reconst} Let $X=G/K$ be a noncompact symmetric space defined as above.
Let $\sigma \in \Sch'(\RR), \rho \in \Sch(\RR)$. 
Then,
\begin{align*}
    &S[R[f;\rho]](x) \notag \\
    &= \int_{\lieA^*\times\bdX\times\RR} R[f;\rho](a,u,b)\sigma(a\iprod{x,u}-b) 
    e^{\varrho\iprod{x,u}}\dd a \dd u \dd b \notag \\
    &= \iiprod{\sigma,\rho} f(x),
\end{align*}
where the equality holds at every point $x \in X$ when $f \in C_c^\infty(X)$, and in $L^2$ when $f \in L^2(X)$.
\end{thm}
The proof is given in \refapp{proof.reconst}, which is parallel to \refsec{solve}. 

As a result, while the Euclidean ridgelet transform is a scalar product of function $f(\xx)$ and co-feature map $\rho(\aa\cdot\xx-b)$,
we revealed that the ridgelet transform on a symmetric space $X$ is a scalar product  of function $f$ and co-feature map $\rho( a\iprod{x,u}-b )$
\emph{with} auxiliary weights $e^{\varrho\iprod{x,u}}$ in the input data domain $X$ and $|\cc(\lambda)|^{-2}$ in the Fourier domain $\lieA^*\times\bdX$.
In geometric deep learning, it has been an open question how to naturally formulate the \emph{affine map} $\aa\cdot\xx-b$ and \emph{element-wise activation} $\sigma$ for  each point $x$ on a manifold \emph{without} depending on the specific choice of coordinates. From the perspective of harmonic analysis on the symmetric space, our answer is to embed the data $x \in X$ into the flat space $\lieA=\RR^r$ via the vector-valued composite distance $\iprod{x,u}$. %

\subsection{$cc$-Universality}
By discretizing the reconstruction formula $S[R[f;\rho]]=f$, 
we can construct a finite network $f_n$ that approximates an arbitrary given function $f$.
This is the primitive idea behind the constructive proof of the following $cc$-universality.

Let $\Delta_\theta^n$ be a forward difference operator with difference $\theta>0$, defined by
    $\Delta_\theta^{1}[\sigma](t) := \sigma(t + \theta) - \sigma(t)$ and %
    $\Delta_\theta^{n+1}[\sigma](t) := \Delta_\theta^{1} \circ \Delta_\theta^{n}[\sigma](t)$. 

\begin{thm}[$cc$-universality of finite networks on symmetric space] \label{thm:cc}
Suppose that 
 there exists $k \ge 0$ and $\theta>0$ such that 
    $\Delta_\theta^k[\sigma] \in L^\infty(\RR)$ and Lipschitz continuous.
Then, the finite neural networks of the form 
\begin{align*}
    f_n(x) = \sum_{i=1}^n c_i \sigma( a_i \iprod{x,u_i} - b_i ) e^{\varrho\iprod{x,u_i}}, \quad x \in X
\end{align*}
are $cc$-universal, that is,
for any compact set $\cdom \subset X$, 
and continuous function $f \in C(\cdom)$, 
there exists a sequence of finite networks such that $\| f_n - f \|_{C(Z)} \to 0$ as $n \to \infty$.
\end{thm}
The proof is given in \refapp{cc.proof}.

\section{Examples: HNNs}%

We instantiate a continuous (horospherical) hyperbolic neural network (HNN) on the Poincar\'e ball model $\Hp^m$. In \refapp{spdnet}, we further instantiate a continuous neural network on the SPD manifold $\spd_m$ (SPDNet).

\subsection{Continuous HNN}%
\begin{dfn} For any $\xx \in \Hp^m$, put
\begin{align*}
    &
    S[\gamma](\xx)
    \\
    &:= \int_{\RR\times\Sph^{m-1}\times\RR} \gamma(a,\uu,b) \sigma(a\iprod{\xx,\uu}-b)
    e^{\varrho\iprod{\xx,\uu}} \dd a \dd\uu \dd b,
\end{align*}
where $\iprod{\xx,\uu} = \log \frac{1-|\xx|^2}{|\xx-\uu|^2}$ for any $(\xx,\uu) \in \Hp^m \times \bdHp^m$.
\end{dfn}
We note that the weight function $\exp(\iprod{\xx,\uu}) = \frac{1-|\xx|^2}{|\xx-\uu|^2}$ is known as the Poisson kernel.

\begin{dfn} For any $(a,\uu,b) \in \RR\times\Sph^{m-1}\times\RR$,
\begin{align*}
    &
    R[f;\rho](a,\uu,b)
    \\
    &= \int_{\Hp^m} \cc[f](\xx)\overline{\rho(a\iprod{\xx,\uu}-b)}e^{\varrho\iprod{\xx,\uu}} \frac{2^m \dd \xx}{(1-|\xx|^2)^m},
\end{align*}
where for any $\xx\in\Hp^m$,
\begin{align*}
        \cc[f](\xx)
    &= \int_{\RR\times\Sph^{m-1}} \widehat{f}(\lambda,\uu) e^{(i\lambda+\varrho)\iprod{\xx,\uu}}  \frac{\dd\lambda\dd\uu}{|W||\cc(\lambda)|^{4}}.
\end{align*}
\end{dfn}

As a consequence of the general results, the following reconstruction formula holds.
\begin{cor} For any $\sigma \in \Sch'(\RR), \rho \in \Sch(\RR)$,
\begin{align*}
    S[R[f;\rho]](\xx) = \iiprod{\sigma,\rho} f(\xx),
\end{align*}
where
\begin{align*}
    \iiprod{\sigma,\rho} := \frac{1}{2\pi}\int_{\RR} \sigma^\sharp(\omega)\overline{\rho^\sharp(\omega)}|\omega|^{-1}\dd\omega,
\end{align*}
where the equality holds at every point $x \in \Hp^m$ when $f \in C_c^\infty(\Hp^m)$, and in $L^2$ when $f \in L^2(\Hp^m)$.
\end{cor}

\section{Discussion}

We have devised the fully-connected layer on noncompact symmetric space $X=G/K$, and presented the closed-form expression of the ridgelet transform. The reconstruction formula $S[R[f]]=f$ is further applied to present a constructive proof of the $cc$-universality of finite fully-connected networks on $X$. 
This is the first universality result that covers a wide range of space $X$ and activation functions $\sigma$, associated with a constructive proof in a unified manner. In fact, we do not need to restrict $X$ to be the hyperbolic space or the SPD manifold, nor need to restrict $\sigma$ to be ReLU.

Parallel to the Euclidean case explained in \refsec{geometric.fc.e}, the fully-connected layer $\sigma(a\iprod{x,u}-b)$ on $X$ can also be understood as a wavelet function on a composite distance $d(x,\xi)$ from the point $x$ to a horosphere $\xi$. To see this, we use the fact that a set $\xi(x,u) := \{ y \in X \mid \iprod{x,u}=\iprod{y,u} \}$ is a horosphere through point $x$ with normal $u$.
Given $a,b$ and $u$, put $\xi' := \{ y \in X \mid a \iprod{y,u} = b\}$. Then, for an arbitrary base point $y \in \xi'$, the horosphere $\xi(y,u)$ is a subset of $\xi'$.
Following the notations in \reffig{poincare}, 
suppose $u = kM$, 
and let $x_0 \in \xi(x,u)$ and $y_0 \in \xi'$ be points satisfying $x = k a_x[o]$ and $y_0 = k a_\xi[o]$ for some $a_x, a_\xi \in A$ respectively. Then, $\iprod{x_0,u} - \iprod{y_0,u} = d(x_0,y_0)$, and thus we have
\begin{align*}
\sigma(a \iprod{x,u} - b)
&= \sigma( a d(x_0, y_0)  ) %
= \sigma( a d(x, \xi(y_0,u) )  ). %
\end{align*}
The ordinary wavelet transform can detect/localize a singularity at a point in a signal \citep[see, e.g.,][]{Mallat2009book}, such as the singularity of signal $f(t)=1/|t|$ at the origin $t=0$.
Hence, a wavelet on a distance $d(x,\xi)$ turns out to be a detector of a sigularity along a horosphere $\xi$.

Based on this coordinate-free reformulation, 
given a family $\Xi$ of geometric objects $\xi \subset X$,
we can devise a fully-connected layer on \emph{an arbitrary metric space} $X$ as
\begin{align*}
    S[\gamma](x) := \int_{\RR\times\Xi} \gamma(a,\xi) \sigma( a d(x,\xi) ) \dd a \dd \xi.
\end{align*}
If we have a nice coordinates such as $(s,t) \in \RR^m \times \RR^m$ satisfying $d(x(t),\xi(s)) = t-s$, then we can turn it to the Fourier expression and hopefully obtain the ridgelet transform.

\paragraph{Comparison to HNNs.} While \citet{Wang2021} and we employed a horosphere as the geometric object $\xi$, the original HNNs \citep{Ganea2018hnn,Shimizu2021}
employed not a horosphere but a set $\xi_{geo.}(\xx,\uu)$ of geodesics perpendicular to a normal vector $\uu \in \Sph^{m-1} (\subset T_{\xx} \BB^m)$ at a point $\xx \in \BB^m$, called the \emph{Poincar\'e hyperplane}. Since a Euclidean hyperplane can be understood as a set of Euclidean geodesics as well as a Euclidean sphere with infinite radius, both the Poincar\'e hyperplane and horosphere can be regarded as a hyperbolic counterpart of the Euclidean hyperplane. In fact, there are two types of Radon transform on the hyperbolic space: geodesic and horospherical Radon transforms. We conjecture that both networks can be understood as wavelet analysis on the Radon domain, but the original HNNs are based on the geodesic Radon transform, while ours are based on the horospherical Radon transform.

One of our reviewers kindly notified us that \citet{Yu2022} introduced the same weight function $\exp(\iprod{\xx,\uu})$, or the Poisson kernel, in graph learning by investigating the hyperbolic Laplacian. This is not a coincidence since the Helgason-Fourier transform decomposes function by the eigenfunctions of the Laplace-Beltrami operator on $X$.

\paragraph{Comparison to SPDNets.}
In a higher rank symmetric space, the Riemannian distance is not a complete two-point invariance, but the vector-valued distance is \citep[see, e.g.,][]{Kapovich2017}. 
\citet{Lopez2021} have recently utilized it. The original SPDNets \citep{Huang2017,Dong2017,Gao2019,Brooks2019,Brooks2019a} are composed of the BiMap layer $x \mapsto w^\top x w$ for $x \in \spd_m$ with an orthonormal projection matrix $w \in \RR^{m \times k}$ satisfying $w^\top w = I$, which extends the scalar product, and the ReEig layer $x \mapsto u^\top \max(0, \lambda - b) u$ via the spectral decomposition $x = u^\top \lambda u$, which extends the pointwise activation with ReLU. While we applied nonlinear activation $\sigma$ on $\log \lambda(x) \in \lieA^* \cong \RR^r$, the original ReEig layer applied $\sigma$ on $\lambda(x) \in A \cong \RR^r_+$. It would be a routine to modify our main results to the original formulations.

\section*{Acknowledgements}
The authors are grateful to anonymous reviewers for their valuable comments.
This work was supported by JSPS KAKENHI 18K18113, JST CREST JPMJCR2015 and JPMJCR1913, JST PRESTO JPMJPR2125, and JST ACT-X JPMJAX2004.

\bibliography{libraryS}
\bibliographystyle{icml2022}

\newpage
\appendix
\onecolumn

\section{Poincar\'{e} Disk $D$ as Noncompact Riemannian Symmetric Space} \label{sec:poincare.ss}
Following \citet[Intro.~\S~4]{Helgason.GGA}, we review the homogeneous space aspect of a hyperbolic space. Note that the Riemannian metric here drops the factor $\times 2^2$.
Let $D := \{ z \in \CC \mid |z| < 1 \}$ be the unit open disk in $\CC$ equipped with the Riemannian metric $g_z(u,v) = \eprod{u,v}/(1-|z|^2)^2$ for any tangent vectors $u,v \in T_z D$ at $z \in D$, where $\eprod{\cdot,\cdot}$ denotes the Euclidean inner product in $\RR^2$.
Let $\bdD := \{ u \in \CC \mid |u|=1\}$ be the boundary of $D$ equipped with the uniform probability measure $\dd u$.
Namely, $D$ is the \emph{Poincar\'e disk model of hyperbolic plane $\HH^2$}.
On this model, the Poincar\'e metric between two points $z,w \in D$ is given by $d(z,w)=\tanh^{-1} | (z-w)/(1-zw^*) |$, and the volume element is given by $\dd z = (1-(x^2+y^2))^{-2}\dd x \dd y$.

Consider now the group
\begin{align*}
G = SU(1,1)
:= \left\{ \begin{pmatrix} \alpha & \beta \\ \beta^* & \alpha^* \end{pmatrix} \Bigg| (\alpha,\beta)\in\CC^2, |\alpha|^2-|\beta|^2=1 \right\},
\end{align*}
which acts on $D$ (and $\bdD$) by
\begin{align*}
    g \cdot z := \frac{\alpha z+\beta}{\beta^*z+\alpha^*}, \quad z \in D \cup \bdD.
\end{align*}
The $G$-action is transitive, conformal, and maps circles, lines, and the boundary into circles, lines, and the boundary.
In addition, consider the subgroups 
\begin{align*}
    K &:= SO(2) = \left\{ k_\phi := \begin{pmatrix} e^{i \phi} & 0 \\ 0 & e^{-i\phi} \end{pmatrix} \Bigg| \phi \in [0,2\pi) \right\}, \\
    A  &:= \left\{ a_t := \begin{pmatrix} \cosh t & \sinh t \\ \sinh t & \cosh t \end{pmatrix} \Bigg| t \in \RR \right\}, \\
    N  &:= \left\{ n_s := \begin{pmatrix} 1+is & -is \\ is & 1-is \end{pmatrix} \Bigg| s \in \RR \right\}, \\
    M &:= C_K(A) = \left\{ k_0 = \begin{pmatrix} 1 & 0 \\ 0 & 1 \end{pmatrix}, k_\pi = \begin{pmatrix} -1 & 0 \\ 0 & -1 \end{pmatrix} \right\}
\end{align*}
The subgroup $K := SO(2)$ fixes the origin $o \in D$. So we have the identifications
\begin{align*}
    D = G/K = SU(1,1)/SO(2), \quad \mbox{and} \quad \bdD = K/M = \Sph^{1}.
\end{align*}

On this model, the following are known 
(1) that $m=\dim \lieA=1$, $|W|=1$, $\varrho=1$, and $|\cc(\lambda)|^{-2} = \frac{\pi \lambda}{2}\tanh (\frac{\pi \lambda}{2})$ for $\lambda \in \lieA^* = \RR$,
(2) that the geodesics are the circular arcs perpendicular to the boundary $\bdD$, and (3) that the horocycles are the circles tangent to the boundary $\bdD$. Hence, let $\xi(x,u)$ denote the horocycle $\xi$ through $x \in D$ and tangent to the boundary at $u \in \bdD$; and let $\iprod{x,u}$ denote the signed distance from the origin $o \in D$ to the horocycle $\xi(x,u)$.

In order to compute the distance $\iprod{z,u}$, we use the following fact: 
The distance from the origin $o$ to a point $z=r e^{iu}$ is $d(o,z) = \tanh^{-1}| (0-z)/(1-0z^*) | = \frac{1}{2}\log\frac{1+r}{1-r}$. Hence, let $c \in D$ be the center of the horocycle $\xi(z,u)$, and let $w \in D$ be the closest point on the horocycle $\xi(z,u)$ to the origin. By definition, $\iprod{z,u} = d(o,w)$. But we can find the $w$ via the cosine rule:
\begin{align*}
    \cos zou = \frac{|u|^2+|z|^2-|z-u|^2}{2|u||z|} = \cos zoc = \frac{|z|^2 + |\frac{1}{2}(1+|w|)|^2 - |\frac{1}{2}(1-|w|)|^2}{2|z||\frac{1}{2}(1+|w|)|},
\end{align*}
which yields the tractable formula:
\begin{align*}
\iprod{z,u} = \frac{1}{2} \log \frac{1+|w|}{1-|w|} = \frac{1}{2} \log \frac{1-|z|^2}{|z-u|^2}, \quad (z,u) \in D \times \bdD.
\end{align*}

\newpage
\section{Proofs}
\subsection{\refthm{reconst} (Reconstruction Formula)} \label{sec:proof.reconst}
\begin{proof}
We identify the scale parameter $a \in \lieA^*$ with vector $\aa \in \RR^r$.
\paragraph{Step~1.} Since $b \in \RR$, the Fourier expression is given by
\begin{align*}
    S[\gamma](x)
    &:= \int_{\RR^r\times\bdX\times\RR} \gamma(\aa,u,b)\sigma(\aa\cdot\iprod{x,u}-b)e^{\varrho\iprod{x,u}}\dd\aa\dd u\dd b \notag \\
    &= \frac{1}{2\pi}\int_{\RR^r\times\bdX\times\RR} \gamma^\sharp(\aa,u,\omega)\sigma^\sharp(\omega)e^{(i\omega\aa+\varrho)\iprod{x,u}}\dd\aa\dd u\dd\omega.
\end{align*}

\paragraph{Step~2.}
By changing the variables as $(\aa,\omega) = (\llambda/\omega,\omega)$ with $\dd\aa\dd\omega = |\omega|^{-r}\dd\llambda\dd\omega$, and identifying the vector $\llambda=(\lambda_1,\ldots,\lambda_r) \in \RR^r$ with $\lambda \in \lieA^*$, we have
\begin{align*}
    S[\gamma](x)
    &= \frac{1}{2\pi}\int_{\RR} \left[ \int_{\lieA^*\times\bdX} \gamma^\sharp(\lambda/\omega,u,\omega) e^{(i\lambda+\varrho)\iprod{x,u}} \dd\lambda\dd u\right] \sigma^\sharp(\omega)|\omega|^{-r}\dd\omega.
\end{align*}

\paragraph{Step~3.}
Since inside the bracket $[\cdots]$ is the inverse Helgason-Fourier transform (excluding the Plancherel measure $|\cc(\lambda)|^{-2}$), put a separation-of-variables form as
\begin{align*}
    \gamma_{f,\rho}^\sharp(\lambda/\omega,u,\omega) = \widehat{f}(\lambda,u)\overline{\rho^\sharp(\omega)}|\cc(\lambda)|^{-2},
\end{align*}
we have a particular solution:
\begin{align*}
    S[\gamma_{f,\rho}](x)
    &= \left(\frac{|W|}{2\pi}\int_{\RR} \sigma^\sharp(\omega)\overline{\rho^\sharp(\omega)}|\omega|^{-r}\dd\omega \right)\left(\int_{\lieA^*\times\bdX} \widehat{f}(\lambda,u) e^{(i\lambda+\varrho)\iprod{x,u}} \frac{\dd\lambda\dd u}{|W||\cc(\lambda)|^2} \right)
    = \iiprod{\sigma,\rho} f(x),
\end{align*}
where we put
\begin{align*}
    \iiprod{\sigma,\rho} := \frac{|W|}{2\pi} \int_{\RR} \sigma^\sharp(\omega)\overline{\rho^\sharp(\omega)}|\omega|^{-r}\dd\omega.
\end{align*}
Here, the equality holds for every point $x \in X$ when $f \in C_c^\infty(X)$, and in $L^2$ when $f \in L^2(X)$.

In particular, the ridgelet transform is calculated as
\begin{align*}
    R[f;\rho](\aa,u,b)
    &:= 
    \frac{1}{2\pi}\int_{\RR}
    \gamma_{f,\rho}^\sharp(\aa,u,\omega)
    e^{i\omega b} \dd\omega \notag \\
    &= \frac{1}{2\pi}\int_{\RR} \widehat{f}(\omega\aa,u)|\cc(\omega\aa)|^{-2}\overline{\rho^\sharp(\omega)}e^{i\omega b}\dd\omega \notag \\
    &= \frac{1}{2\pi}\int_{\RR\times X} \cc[f](x) \overline{\rho^\sharp(\omega)} e^{(-i\omega\aa+\varrho)\iprod{x,u}+i\omega b}\dd x\dd\omega \notag \\
    &= \int_{X} \cc[f](x)\overline{\rho(\aa\cdot\iprod{x,u}-b)} e^{\varrho\iprod{x,u}}\dd x,
\end{align*}
where we put $\cc[f]$ as a Helgason-Fourier multiplier satisfying $\widehat{\cc[f]}(\lambda,u) = \widehat{f}(\lambda,u)|\cc(\lambda)|^{-2}$. 
\end{proof}
\subsection{\refthm{cc} ($cc$-Universality)} \label{sec:cc.proof}
\newcommand{\fa}{{f_c}}
\newcommand{\fb}{{f_V}}
\newcommand{\fn}{{f_n}}

\paragraph{Additional Notation.}
For a function $f$ on a set $X$, $\| f \|_{C(X)} := \sup_{x \in X} | f(x) |$ denotes the uniform norm on $X$.

For any integer $d >0$ and vector $ \vv \in \RR^d$, $|\vv|$ denotes the Euclidean norm, and $\iprod{\vv} := \sqrt{1 + |\vv|^2}$. For any positive number $t > 0$, $\triangle^{t/2}$ and $\iprod{\triangle}^t$ denote fractional differential operators defined as Fourier multipliers: for any $\phi \in \Sch'(\RR^d)$,
\begin{align*}
\triangle^{t/2}[\phi](\vv) := \frac{1}{(2\pi)^d} \int_{\RR^d} |\uu|^t \widehat{\phi}(\uu) e^{i\uu\cdot\vv}\dd\uu, \quad 
\iprod{\triangle}^{t/2}[\phi](\vv) := \frac{1}{(2\pi)^d} \int_{\RR^d} (1+|\uu|^2)^{t/2} \widehat{\phi}(\uu) e^{i\uu\cdot\vv}\dd\uu.
\end{align*}
In particular when $t=2$, $\triangle^{t/2}$ coincides with the ordinary Laplacian on $\RR^d$. 

\begin{proof} 
We will show that for any compact set $\cdom \subset X$, positive number $\eps > 0$, compactly-supported continuous function $f \in C(\cdom)$, there exists a finite network $f_n$ such that $\| f - f_n \|_{C(\cdom)} < \eps$.

Since $\sum_{i=1}^n c_i \Delta_\theta^k[\sigma]( a_i \iprod{x,u_i} - b_i ) e^{\varrho\iprod{x,u_i}}$ is rewritten as another finite model $\sum_{i=1}^{n'} c_i' \sigma( a_i' \iprod{x,u_i'} - b_i' ) e^{\varrho\iprod{x,u_i'}}$, it suffice to consider the case $k=0$. In the following, we assume that $\sigma (= \Delta_\theta^0[\sigma])$ is bounded and Lipschitz continuous. So, put $M_\sigma := \| \sigma \|_{L^\infty(\RR)}$ and $L_\sigma := \lip(\sigma)$. As a consequence of the Iwasawa decomposition, the composite distance $\iprod{x,u}$ is $C^\infty$-smooth and thus Lipschitz continuous. Hence, put $L_c := \sup_{x \in \cdom} \sup_{u,u' \in \bdX} |\iprod{x,u} - \iprod{x,u'}|/d(u,u')$,
$L_{e} := \sup_{x \in \cdom} \sup_{u,u' \in \bdX} |\exp(\varrho\iprod{x,u}) - \exp(\varrho\iprod{x,u'})|/d(u,u')$,
and $M_e := \sup_{x \in \cdom, u \in \bdX} | \exp(\varrho\iprod{x,u}) |$.

\paragraph{Step~1 ($f \sim \fa$).}
By the density of $C_c^\infty(X)$ in $C(\cdom)$ with respect to the uniform norm, we can take a compactly-supported smooth function $\fa \in C_c^\infty(\cdom)$ satisfying $\| f - \fa \|_{C(\cdom)} < \eps/3$. Since $\fa$ is sufficiently smooth and integrable, there exists a compactly-supported smooth function $\rho \in C_c^\infty(\RR)$ such that 
\begin{align*}
    S[R[\fa;\rho]](x) = \fa(x) \mbox{ at every point } x \in X.
\end{align*}
For example, 
take a compactly-supported smooth function $\rho_0 \in C_c^\infty(\RR)$, and put $\rho(b) := \triangle_b^{r/2}[\rho_0](b) = \frac{1}{2\pi}\int_\RR |\omega|^r \rho_0^\sharp(\omega) e^{ib\omega}\dd\omega$.
Then, $\iiprod{\sigma,\rho} = \frac{|W|}{2\pi}\int_{\RR} \sigma^\sharp(\omega) \overline{\rho^\sharp(\omega)}|\omega|^{-r}\dd \omega = \frac{|W|}{2\pi}\int_{\RR} \sigma^\sharp(\omega) \overline{\rho_0^\sharp(\omega)} \dd \omega = |W| \int_{\RR} \sigma(b) \overline{\rho_0(b)} \dd b = |W|\iprod{\sigma,\rho_0}_{L^2(\RR)}$, which is an ordinary functional inner product, and it is easy to find a $\rho_0$ satisfying $\iprod{\sigma,\rho_0}_{L^2(\RR)} \neq 0$. By normalizing $\rho' := \rho/\iiprod{\sigma,\rho}$, we can find the $\rho'$. We refer to \citet{Sonoda2015acha} and \citet{Sonoda2021ghost} for more details on the scalar product $\iiprod{\sigma,\rho}$.

\paragraph{Step~2 ($R[\fa;\rho]$).}
To show a discretization $\fn$ of the reconstruction formula converges to $\fa$ in $C(\cdom)$, 
it is convenient to regard 
the integrand %
\begin{align*}
    \phi(a,u,b)(x) := R[\fa;\rho](a,u,b)\sigma(a\iprod{x,u}-b)e^{\varrho\iprod{x,u}}
\end{align*}
as a vector-valued function $\phi:\lieA^*\times\bdX\times\RR \to C(\cdom)$,
and the integration $S[\gamma](x) = \int_{\lieA^*\times\bdX\times\RR} \phi(a,u,b)(x) \dd a \dd u \dd b$ as a Bochner integral.
Since $\fa$ is $C^\infty$-smooth, $R[\fa;\rho](a,u,b)$ is bounded and decays rapidly in $a$, and thus $\phi$ is Bochner integrable, that is, 
\begin{align*}
    \int_{\lieA^*\times\bdX\times\RR} \| \phi(a,u,b) \|_{C(\cdom)} \dd a \dd u \dd b < \infty.
\end{align*}
To see this, 
the decay property is estimated as follows. For any positive numbers $s,t>1$,
\begin{align*}
| R[\fa;\rho](a,u,b) |
&= \frac{1}{2\pi} \Bigg| \int_{\RR} \widehat{\fa}(\omega a,u)|\cc(\omega a)|^{-2} \overline{\rho^\sharp(\omega)} e^{i\omega b}\dd\omega \Bigg| \notag\\
&= \frac{1}{2\pi} \Bigg| \int_{\RR} \iprod{\omega a}^s \iprod{\omega a}^{-s} \iprod{b}^{t} \iprod{b}^{-t} \widehat{\fa}(\omega a,u)|\cc(\omega a)|^{-2}  \overline{\rho^\sharp(\omega)}  e^{i\omega b}\dd\omega \Bigg| \notag \\
&\le \frac{1}{2\pi} \Bigg| \int_{\RR} \iprod{\omega a}^s    \widehat{\fa}(\omega a,u)|\cc(\omega a)|^{-2}  \iprod{\omega}^{-s} \overline{\rho^\sharp(\omega)} \iprod{\triangle_\omega}^{t} e^{i\omega b}\dd\omega \Bigg| \iprod{a}^{-s} \iprod{b}^{-t}, %
\end{align*}
which asserts the integrability as below
\begin{align*}
    \int_{\lieA^*\times\bdX\times\RR} \| \phi(a,u,b) \|_{C(\cdom)} \dd a \dd u \dd b 
    &\le M_\sigma M_e \| R[f_c;\rho] \|_{L^1(X)}
    \lesssim \int_{\lieA^*\times\bdX\times\RR} \iprod{a}^{-s} \iprod{b}^{-t} \dd a \dd u \dd b < \infty.
\end{align*}

\paragraph{Step~3 ($\fa \sim \fb \sim \fn$).}
Next, take a compact domain $V := \{ (a,u,b) \in \lieA^*\times\bdX\times\RR \mid |a_i| \le \delta/2, |b| \le \delta/2 \}$, namely 
the product of an $(r+1)$-dimensional hypercube and the compact manifold $\bdX$,
and put a band-limited function
\begin{align*}
    \fb(x) 
    := \int_V \phi(a,u,b)(x) \dd a \dd u \dd b,
\end{align*}
so that $\| \fa - \fb \|_{C(\cdom)} < \eps/3$ (by letting $\delta$ sufficiently large).
For each $n \in \NN$, let $V = \bigsqcup_{i \in I_n} V_{ni}$ be a disjoint decomposition of $V$ into a disjoint family of $|I_n|$ subsets $V_{ni}$ with diameter at most $d_n = O(1/n)$. Since $V$ is a compact manifold, each volume $\vol(V_{ni})$ decays at $O(n^{-{\dim V}})$ as $n \to \infty$, and the cardinality $|I_n|$ ($\approx$ $d_n$-covering number) grows at the reciprocal $O(n^{\dim V})$.
From each subset $V_{ni}$, take a point $(a_{ni},u_{ni},b_{ni})$ satisfying 
\begin{align*}
c_{ni} := \int_{V_{ni}} R[\fa;\rho](a,u,b) \dd a \dd u \dd b = R[\fa;\rho](a_{ni},u_{ni},b_{ni}) \vol(V_{ni}),
\end{align*}
and put a finite network as
\begin{align*}
    \fn(x) := \sum_{i \in I_n} c_{ni} \sigma(a_{ni}\iprod{x,u_{ni}}-b_{ni})e^{\varrho\iprod{x,u_{ni}}}.
\end{align*}
In addition, we use
\begin{align*}
    \phi_{ni}(x) := \phi(a_{ni},u_{ni},b_{ni})(x), \quad \mbox{and} \quad
    \phi_n(a,u,b)(x) := \sum_{i \in I_n} \ind_{V_{ni}}(a,u,b) \phi_{ni}(x),
\end{align*}
so that 
\begin{align*}
\fn(x) = \sum_{i \in I_n} \phi_{ni}(x) \vol(V_{ni}) = \int_V \phi_n(a,u,b)(x)\dd a \dd u \dd b.
\end{align*}

\paragraph{Step~4 ($\fb \sim \fn$).}
We show $\fn \to \fb$ in $C(\cdom)$. Put $M_R := \| R[\fa;\rho] \|_{C(V)}$ and $L_R := \lip(R[\fa;\rho])$.
For every $n \in \NN$, since
\begin{align*}
    \| \fb - \fn \|_{C(\cdom)}
    &= \sup_{x \in \cdom} \Bigg| \int_V \phi(a,u,b)(x) \dd a \dd u \dd b - \int_V \phi_n(a,u,b)(x) \dd a \dd u \dd b\Bigg| \\
    &\le \int_V  \| \phi(a,u,b) - \phi_n(a,u,b) \|_{C(\cdom)} \dd a \dd u \dd b,
\end{align*}
it suffice to show that (1) $\phi_n$ is a.e. dominated by an integrable function, and (2) converges a.e. to $\phi$.
In the following, we fix an arbitrary $(a,u,b) \in V_{ni}$. First, 
$\phi_n$ is uniformly dominated by a constant function, which is in $L^1(V)$, that is,
\begin{align*}
    \| \phi_n(a,u,x) \|_{C(\cdom)} = \| \phi_{ni} \|_{C(\cdom)} \le \sup_{(a',u',b') \in V_{ni}} \| \phi(a',u',b') \|_{C(\cdom)} \le M_R  M_\sigma M_e.
\end{align*}
Second, $\phi_n$ coverges to $\phi$ a.e.:
\begin{align*}
&\| \phi(a,u,b) - \phi_n(a,u,b) \|_{C(\cdom)}\\
&= \| \phi(a,u,b) - \phi_{ni} \|_{C(\cdom)} \\
&= \sup_{x \in \cdom} \Big| R[\fa;\rho](a,u,b)\sigma(a\iprod{x,u}-b)e^{\varrho\iprod{x,u}} - R[\fa;\rho](a_{ni},u_{ni},b_{ni})\sigma(a_{ni}\iprod{x,u_{ni}}-b_{ni}) e^{\varrho\iprod{x,u_{ni}}} \Big| \\ 
&\le \sup_{x \in \cdom} \Big| R[\fa;\rho](a,u,b) \Big| \Big|\sigma(a\iprod{x,u}-b)e^{\varrho\iprod{x,u}} - \sigma(a_{ni}\iprod{x,u_{ni}}-b_{ni}) e^{\varrho\iprod{x,u_{ni}}} \Big| \notag \\
&\qquad + \sup_{x \in \cdom} \Big| R[\fa;\rho](a_{ni},u_{ni},b_{ni}) - R[\fa;\rho](a,u,b) \Big| \Big| \sigma(a_{ni}\iprod{x,u_{ni}}-b_{ni}) e^{\varrho\iprod{x,u_{ni}}} \Big| \\
&\le M_R \left( L_\sigma M_e \sup_{x \in \cdom} \Big| a\iprod{x,u} - a_{ni}\iprod{x,u_{ni}} + (b-b_{ni})\Big| + M_\sigma L_e d(u,u_{ni}) \right) \notag\\
&\qquad+ 
M_\sigma M_e L_R \Big|d( (a,u,b), (a_{ni},u_{ni},b_{ni}) )\Big| \notag \\
&\lesssim d_n = O(1/n) \to 0, \quad n \to \infty.
\end{align*}
Therefore, the dominated convergence theorem for the Bochner integral yields
\begin{align*}
    \| \fb - \fn \|_{C(\cdom)}
    &\le \int_V \| \phi(a,u,b) - \phi_n(a,u,b) \|_{C(\cdom)} \dd a \dd u \dd b
    \to 0, \quad n \to \infty.
\end{align*}
Hence by letting $n$ sufficiently large, we have $\| \fn - \fb \|_{C(\cdom)} < \eps/3$. 

To sum up, we have shown the $cc$-universality:
\begin{align*}
    \| f - \fn \|_{C(\cdom)} \le \| f - \fa \|_{C(\cdom)} + \| \fa - \fb \|_{C(\cdom)} + \| \fb - \fn \|_{C(\cdom)} < \eps. %
\end{align*}
\end{proof}

\newpage
\section{Further Examples: SPDNets} \label{sec:spdnet}
\subsection{SPD Manifold}%
Following \citet[][Chapter~1]{Terras2016}, we introduce the SPD manifold.
On the space $\spd_m$ of $m \times m$ symmetric positive definite (SPD) matrices, the Riemannian metric is given by
\begin{align*}
    \mathfrak{g}_{x} := \tr\left( (x^{-1} \dd x)^2 \right), \quad x \in \spd_m
\end{align*}
where $x$ and $\dd x$ denote the matrices of entries $x_{ij}$ and $\dd x_{ij}$.

Put $G=GL(m,\RR)$, then the Iwasawa decomposition $G=KAN$ is given by $K=O(m), A=D_+(m), N=T_1(m)$; and the centralizer $M = C_K(A)$ is given by $M=D_{\pm 1}$ (diagonal matrices with entries $\pm 1$). The quotient space $G/K$ is identified with the SPD manifold $\spd_m$ via a diffeomorphism onto, $gK \mapsto g g^\top$ for any $g \in G$; and $K/M$ is identified with the boundary $\bdspd_m$, another manifold of all \emph{singular positive semidefinite} matrices. The action of $G$ on $\spd_m$ is given by $g[x] := g x g^\top$ for any $g \in G$ and $x \in \spd_m$. In particular, the metric $\mathfrak{g}$ is $G$-invariant. According to the \emph{spectral decomposition}, for any $x \in \spd_m$, there uniquely exist $k \in K$ and $a \in A$ such that $x = k[a]$; and according to the \emph{Cholesky (or Iwasawa) decomposition}, there exist $n \in N$ and $a \in A$ such that $x = n[a]$. 

When $x = k[\exp(H)] = \exp(k[H])$ for some $H \in \lieA = D(m)$ and $k \in K$, then the geodesic segment $y$ from the origin $o=I$ (the identity matrix) to $x$ is given by 
\begin{align*}
y(t) = \exp(t k[H]), \quad t \in [0,1]
\end{align*}
satisfying $y(0) = o$ and $y(1) = x$;
and the Riemannian length of $y$ (i.e., the Riemannian distance from $o$ to $x$) is given by $d(o,x) = |H|_E$. So, $H \in \lieA$ is the \emph{vector-valued distance} from $o$ to $x = k[\exp(H)]$.

The $G$-invariant measures are given by $\dd g = |\det g|^{-m} \bigwedge_{i,j} \dd g_{ij}$ on $G$, $\dd k$ to be the uniform probability measure on $K$, $\dd a = \bigwedge_i \dd a_i/a_i$ on $A$, $\dd n = \bigwedge_{1 < i < j \le m} \dd n_{ij}$ on $N$,
\begin{align*}
\dd \mu(x) 
&= |\det x|^{-\frac{m+1}{2}} \bigwedge_{1 \le i \le j \le m} \dd x_{ij} \quad \mbox{on} \quad \spd_m, \notag \\
&= c_m \prod_{j=1}^m a_j^{-\frac{m-1}{2}} \prod_{1 \le i < j \le m} |a_i - a_j| \dd a \dd k,
\end{align*}
where the second expression is for the polar coordinates $x \gets k[a]$ with $(k,a) \in K \times A$ and $c_m := \pi^{(m^2+m)/4} \prod_{j=1}^m j^{-1} \Gamma^{-1}(j/2)$,
and $\dd u$ to be the uniform probability measure on $\bdspd_m := K/M$.

The vector-valued composite distance from the origin $o$ to a horosphere $\xi(x,u)$ is calculated as 
\begin{align*}
    \iprod{x = g[o], u=kM} = \frac{1}{2} \log \lambda(k^\top[x]), %
\end{align*}
where $\lambda(y)$ denotes the diagonal vector $\lambda$ in the \emph{Cholesky decomposition} $y = \nu[\lambda] = \nu \lambda \nu^\top$ of $y$ for some $(\nu,\lambda) \in NA$.
\begin{proof}
Since $\iprod{x,kM} := -H(g^{-1}k) = \iprod{k^\top [x],eM}$,
it suffices to consider the case $(x,u)=(g[o],eM)$. Namely, we solve $g^{-1} = kan$ for unknowns $(k,a,n) \in KAN$. (To be preceise, we only need $a$ because $\iprod{x,eM} = - \log a$.) Put the Cholesky decomposition $x = \nu[\lambda] = \nu \lambda \nu^\top$ for some $(\nu,\lambda) \in NA$. Then, $a = \lambda^{-1/2}$ because $x^{-1} = (\nu^{-1})^\top \lambda^{-1} \nu^{-1}$, while $x^{-1} = (g g^\top)^{-1} = n^\top a^2 n$.
\end{proof}

The Helgason-Fourier transform and its inversion formula are given by
\begin{align*}
    &\widehat{f}(\ss,u) = \int_{\spd_m} f(x) \overline{e^{\ss\cdot\iprod{x,u}}} \dd \mu(x),\\
    &f(x) = \omega_m \int_{\Re \ss = \rrho} \int_{\bdspd_m} \widehat{f}(\ss,u) e^{\ss\cdot\iprod{x,u}} \dd u \frac{\dd \ss}{|\cc(\ss)|^2},
\end{align*}
for any $(\ss,u) \in \lieA^*_{\CC} \times O(m)$ (where $\lieA^*_{\CC} = \CC^m$) and $x \in \spd_m$. Here, $\omega_m := \prod_{j=1}^m \frac{\Gamma(j/2)}{j (2\pi i) \pi^{j/2}}$, $\rrho = (-\frac{1}{2},\ldots,-\frac{1}{2},\frac{m-1}{4}) \in \CC^m$, and 
\begin{align*}
    \cc(\ss) = \prod_{1 \le i \le j < m} \frac{B(\frac{1}{2}, s_i + \cdots + s_j + \frac{j-i+1}{2})}{B(\frac{1}{2},\frac{j-i+1}{2})},
\end{align*}
where $B(x,y) := \Gamma(x)\Gamma(y)/\Gamma(x+y)$ is the beta function.

\subsection{Continuous SPDNet}
\begin{dfn} For any $x \in \spd_m$, put
\begin{align*}
    S[\gamma](x)
    &= \int_{\RR^m\times\bdspd_m\times\RR} \gamma(\aa,u,b) \sigma(\aa\cdot\iprod{x,u}-b) e^{\rrho\cdot\iprod{x,u}} \dd \aa \dd u \dd b,
\end{align*}
where for any $(x,u) \in \spd_m \times \bdspd_m$ with $u = kM$ for some $k \in K$,
\begin{align*}
\iprod{x,u} &= \frac{1}{2} \log \lambda(k^\top[x]). %
\end{align*}
\end{dfn}

\begin{dfn} For any $(\aa,b) \in \RR^m\times\RR$,
\begin{align*}
    R[f;\rho](\aa,b)
    &= \int_{\spd_m} \cc[f](x)\overline{\sigma(\aa\cdot\iprod{x,u}-b)} e^{\rrho\cdot\iprod{x,u}} \dd\mu(x),
\end{align*}
where for any $x\in\spd_m$,
\begin{align*}
    \cc[f](x)
    &= \int_{\RR^m\times\bdspd_m} \widehat{f}(i\llambda+\rrho,u) e^{(i\llambda+\rrho)\cdot\iprod{x,u}} \frac{\omega_m\dd\llambda\dd u}{|\cc(i\llambda+\rrho)|^{4}}.
\end{align*}
\end{dfn}

As a consequence of the general results, the following reconstruction formula holds.
\begin{cor} For any $\sigma \in \Sch'(\RR), \rho \in \Sch(\RR)$,
\begin{align*}
    S[R[f;\rho]](x) = \iiprod{\sigma,\rho} f(x),
\end{align*}
where
\begin{align*}
    \iiprod{\sigma,\rho} := \frac{1}{2\pi}\int_{\RR} \sigma^\sharp(\omega)\overline{\rho^\sharp(\omega)}|\omega|^{-m}\dd\omega,
\end{align*}
where the equality holds at every point $x \in \spd_m$ when $f \in C_c^\infty(\spd_m)$, and in $L^2$ when $f \in L^2(\spd_m)$.
\end{cor}

\end{document}